# Collaborative AI Enhances Image Understanding in Materials Science

Ruoyan Avery Yin,[1] Zhichu Ren,[2] Zongyou Yin,[3] Zhen Zhang,[2] So Yeon Kim,[2] Chia-Wei Hsu,[2] Ju Li[2]

[1]Department of Computing Science, National University of Singapore, Singapore
[2]Department of Nuclear Science and Engineering and Department of Materials Science and Engineering, Massachusetts Institute of Technology, Cambridge, MA, USA
[3]Research School of Chemistry, The Australian National University, Canberra, Australian Capital Territory, 2601 Australia

*Abstract* — The Copilot for Real-world Experimental Scientist (CRESt) system empowers researchers to control autonomous laboratories through conversational AI, providing a seamless interface for managing complex experimental workflows. We have enhanced CRESt by integrating a multi-agent collaboration mechanism that utilizes the complementary strengths of the ChatGPT and Gemini models for precise image analysis in materials science. This innovative approach significantly improves the accuracy of experimental outcomes by fostering structured debates between the AI models, which enhances decision-making processes in materials phase analysis. Additionally, to evaluate the generalizability of this approach, we tested it on a quantitative task of counting particles. Here, the collaboration between the AI models also led to improved results, demonstrating the versatility and robustness of this method. By harnessing this dual-AI framework, this approach stands as a pioneering method for enhancing experimental accuracy and efficiency in materials research, with applications extending beyond CRESt to broader scientific experimentation and analysis.

## I. INTRODUCTION

In recent years, the field of image analysis has undergone transformative changes, primarily driven by advances in artificial intelligence (AI). The traditional methods, often manual and time-intensive, are increasingly being augmented or replaced by AI-driven techniques that offer higher accuracy, speed, and efficiency. The integration of sophisticated AI algorithms enables the automatic detection, classification, and interpretation of images, which is particularly crucial in handling large datasets typical of many modern applications. This evolution is pivotal as industries and scientific fields grapple with increasingly complex data that require nuanced analysis beyond human capability alone.

Materials science, a field at the intersection of physics, chemistry, and engineering, is increasingly reliant on advanced imaging techniques for material characterization and analysis. AI's role in this domain is becoming indispensable, particularly for tasks such as phase identification, microstructure analysis, and the development of new materials. By leveraging AI for image analysis, researchers can possibly identify patterns and features that are not discernible by human observers [1] [5], thereby driving innovations in material design, testing, and application. The precision and scalability of AI tools not only enhance research productivity but also pave the way for groundbreaking discoveries that can be translated into practical engineering solutions.

Integration of computer vision (CV) and artificial intelligence (AI) into image analysis operates effectively across various scales, enhancing applications from environmental monitoring to veterinary diagnostics. At the largest scale, AI-driven tools analyze satellite and aerial imagery to manage urban sprawl and track environmental changes. These tools offer real-time monitoring of urban growth and land use, aiding planners in sustainability assessments and informed decisions [6]. At a mid-scale, in veterinary medicine, AI-enhanced imaging techniques are revolutionizing diagnostics by providing rapid and accurate analysis of X-rays, ultrasounds, and MRIs. These technologies can identify abnormalities and patterns that might be missed by humans, improving diagnostic efficiency and treatment outcomes while reducing costs and the need for specialists [4]. However, the micro-world remains relatively unexplored due to technical challenges, yet the potential for AI to make significant contributions exists.

It deserves our attention that recent studies emphasize the synergistic capabilities of collaborative AI models, enhancing decision-making in complex environments through the integration of multiple AI systems, leading to improved performance in tasks that require rapid and comprehensive analysis [2]. Furthermore, the introduction of a multi-agent debate technique for language models demonstrates notable improvements in factual accuracy and reasoning, indicating the potential of collaborative AI to refine cognitive capabilities and enhance performance in intricate tasks like mathematics and chess [2]. These developments showcase the broad applicative scope and effectiveness of collaborative AI strategies.

To drive AI's work, prompt engineering has become crucial in enhancing AI model performance, especially in scenarios that demand precise, context-specific responses. Subtle differences in how prompts are constructed can dramatically affect both the accuracy and relevance of AI outputs [8]. In the realm of medicine, robust prompt engineering was shown to be able to achieve state-of-the-art results on medical benchmarks, outperforming fine-tuning while requiring less computational power and specialized data [3]. These advancements underscore the adaptability and critical role of prompt engineering and its impact on the functionality and applicability of AI technologies across various sectors.

This work investigates the application of AI in microworld image analysis, showcasing its vast untapped potential and versatility across different imaging scales. We utilize two leading AI models: ChatGPT by OpenAI and Gemini by Google. ChatGPT is renowned for its advanced language comprehension and generation capabilities, making it highly effective for managing diverse data inputs and facilitating data integration. Gemini excels in multimodal information processing, adeptly handling data formats such as text, images, audio, and video. Together, these models represent the forefront of AI technologies, with potential synergies that could significantly advance complex analytical tasks, including those in materials science.

This paper explores ChatGPT and Gemini's capabilities in materials science image analysis, focusing on both qualitative and quantitative assessments through collaboration. These models will be used in the application for Copilot for Real-world Experimental Scientist (CRESt) [7]. We aim to investigate how these AI models can be effectively integrated through innovative prompt engineering to improve the accuracy and efficiency of image description and interpretation. Our research systematically analyzes each model's performance, optimizes AI collaboration techniques, and evaluates their effectiveness in real-world materials science applications. To achieve this, our study is structured in two parts: one experiment focuses on a qualitative task, identifying specific material phases within images, while the other experiment focuses on a quantitative task, counting the number of particles present. This approach allows us to comprehensively evaluate the models' performance across different types of tasks, integrating both qualitative and quantitative metrics. This study not only contributes to the academic discourse on multimodal AI applications but also seeks to provide practical insights that can be leveraged in scientific research and industrial practice.

## II. RESULTS

### EXPERIMENT I

In the CRESt application, the large language model (LLM) is used in conjunction with the voice activated web application, where the instruction is given verbally then converted to text. For all our experiments, we will be giving the command "take a picture of the martensite phase with HFW of 80 microns and state the label of the largest ROI when the summarize function is called". Hence this will be the final objective that the LLMs are trying to achieve.

*A. Individual Experiment*

**Experimental design description -** Before ideating the debate mechanism, we first compared the performance of ChatGPT vs Gemini, and possibly invent a novel way to benchmark these two. Hence, we just conducted the experiment for both ChatGPT (CallingGPT version 0.0.1.0) and Gemini (version before 6th Jan 2024), separately.

**Prompt engineering -** The results were not very desirable.
- States an ROI p that is not present in the final scanning electron microscopy (SEM) image generated (Figure S1 in Supplementary Information (SI))
- States an ROI b that does not meet the final objective (Figure S2 in SI)

Hence, we entered the stage of prompt engineering to try to improve the results.

Prompt 1:
- "Information that will be helpful: Martensite phases consist of needle-like structures."
- This change was not kept as this was too specific to the current final objective being tested. Hence, it is not practical to always include relevant information about the final objective when we are generalizing its use case.

Prompt 2:
- "Label can be found in the final image generated."
- This change was kept because after adding this prompt, the final ROI identified was always visible in the final image.

*B. Teamwork Experiment*

We also aimed to explore the potential outcomes of collaborative efforts between ChatGPT and Gemini. Therefore, we initiated a project that allowed these systems to work together toward a common goal. This team-oriented approach was adopted in the anticipation that synergistic collaboration would enhance the overall results.

**Experimental design description**

Definition of keywords:

**'Round'**: The entire process that ends with the final objective being achieved.

**Function call:** In each round, the function image_analysis will be called multiple times. Then the debate will be about what the LLMs observe about the image, what steps to take next, and whether the final objective is achieved.

Steps taken:

1. Let Gemini respond as per normal when the function image_analysis is called.
2. Allow ChatGPT to review Gemini's answer.
   a. If ChatGPT agrees with Gemini's response, the debate ends here.
   b. If ChatGPT disagrees with Gemini's response, ChatGPT will give feedback.
3. Correspondingly, upon receiving ChatGPT's review, Gemini may or may not agree, then refine its answer if necessary.
4. We allow them to review 5 times maximum to maintain efficiency.
5. If an agreement is not reached within 5 times of debate, we arbitrarily take Gemini's response as the final response.

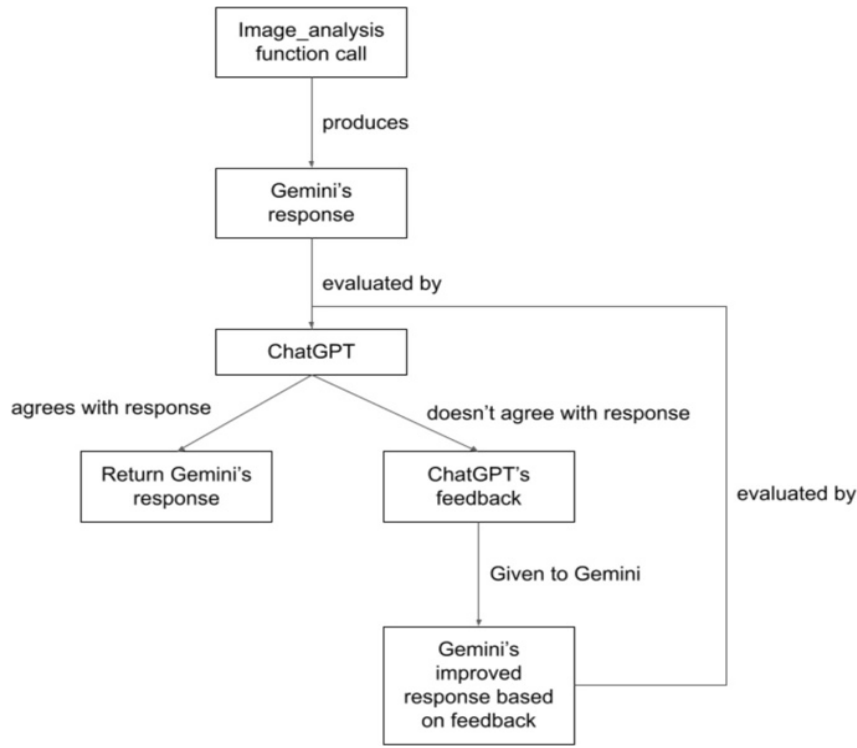

**Figure 1.** AI Model Feedback Loop for Image Analysis. The flowchart depicts the feedback loop between ChatGPT and Gemini in an image analysis task. Gemini produces an initial response, which ChatGPT evaluates. If ChatGPT agrees, the response is returned. If it disagrees, ChatGPT provides feedback for Gemini to refine its response, creating an iterative improvement process.

**Prompt engineering.**

Debate prompts we started with:

Prompt for ChatGPT:

Based on the following analysis, provide your critique or agreement: {gemini_response}. Please collaborate with each other and try to reach an agreement as soon as possible. If you agree, please explicitly state 'I agree'. If you do not agree, please explicitly state 'I do not agree' first. In brackets, state the final objective at the start of your response.

Second prompt for Gemini:

ChatGPT has provided the following critique: {chatgpt_response}. Please collaborate with each other and try to reach an agreement as soon as possible. If you agree, please refine your analysis. If you don't agree, please state why, and repeat your analysis.

Changes in prompt are shown in Table 1.

Note that 'Prompt for ChatGPT' and 'Second prompt for Gemini' will be called in this order repeatedly after Gemini answers for the first time, a maximum of 5 times for each debate.

*C. Image Used and Evaluation*

Examples of SEM images used for region identification and evaluation are shown below, highlighting labeling accuracy and the criteria for correct region of interest (ROI) selection.

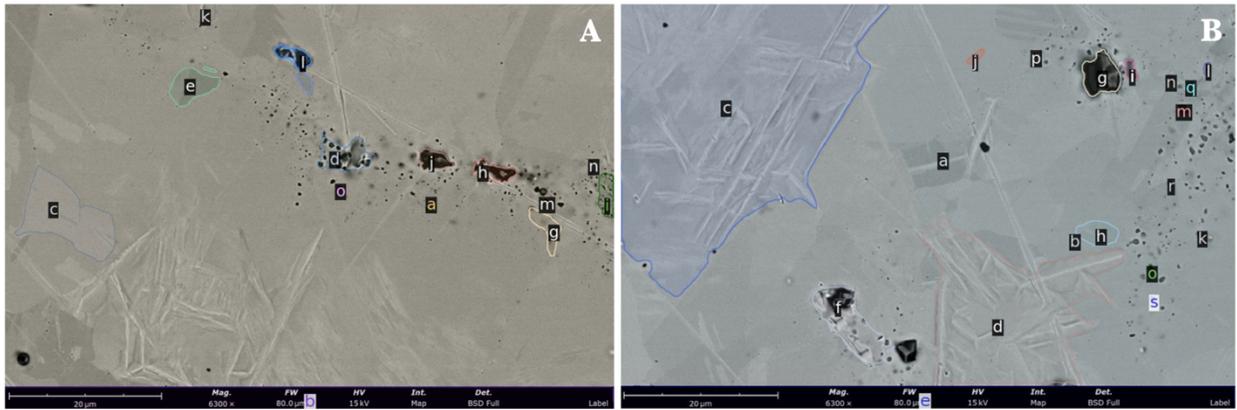

**Figure 2.** Representative examples of region identification and labeling accuracy assessment based on SEM images of samples: (A) In this image, none of the labels correspond to the martensite phase, so any identified conclusion will be marked as incorrect. (B) In this image, the answer will be considered correct if the final region of interest (ROI) identified aligns with regions labeled as a, c, or d.

*D. Script to Facilitate Testing*

A python script was written to facilitate the automation of running 10 rounds, as the whole process can take up to 4 hours. We decided on 10 rounds for cost optimization, due to the costs involved for the tokens sent for the image analysis.

The script was used to provide us with three pieces of information:
1.      Name of the last photo generated
2.      Final Region of Interest (ROI)
3.      Number of function calls

Name of the last photo and final ROI are used for us to manually check the accuracy of the results - whether the martensite phase is indeed visible in the indicated ROI in the corresponding photo. The number of function calls helps us gauge the efficiency of the LLM, where in the case of similar accuracy, the LLM that has the fewer function calls will be the better one as it was able to achieve the same accurate results with less time and resources used.

Prompt added to facilitate the indication of the final ROI: Very important: When the final objective has been achieved, (when the list-summarize function is called), clearly state the label (e.g., a) of the largest ROI identified in this exact format "The final largest ROI is a". Please replace 'a' with the actual result. This sentence must appear in the exact format.

Example output:
20240112_034331 * Number of function calls: 7 * ROI Identified: f.

More examples are shown in Figure S3 of SI.

| Prompt source | Prompts changed | Did it work? | Was this change kept, why or why not. |
|---|---|---|---|
| Gemini | Added:<br>If you don't agree, please state why, and repeat your analysis. | Yes | Yes. There were cases where Gemini just states that it doesn't agree with ChatGPT and does not provide its analysis again. This gives ChatGPT no content to review. |
| System prompt | Added:<br>Please collaborate with each other and try to reach an agreement in 5 rounds of debate. | Yes | Yes, the accuracy improved. |
| System prompt | Added:<br>Keep track of which round of debate you are at, and in the last round, the largest ROI must be stated explicitly. Since there is a debate that occurs for each image analysis, you can get the final largest ROI from the last analysis provided by either ChatGPT and Gemini. | No | No, accuracy worsened. Perhaps keeping track of which round it was at was unnecessary and confusing. |
| System prompt | Added:<br>Please collaborate with each other and try to reach an agreement as soon as possible. | Yes | Yes, accuracy improved to 60%. |
| System prompt | Replaced: Once the final objective has been achieved (after the list-summarize function is called).<br>With: Once the list-summarize function is called, | Yes | Yes, it is more concise to prevent confusing the model. |
| System prompt | Replaced: This sentence must appear in the exact format<br>With: This sentence must appear in the exact format, exactly one time at the end. | Yes | Yes, it is more specific instructions. |
| System prompt | Append the final objective | Yes | Yes, as system prompt holds more importance, adding the final objective to system prompt reminds the model. |

**Table 1.** Prompts and their effectiveness.

*E. INDIVIDUAL PERFORMANCE*

ChatGPT and Gemini was 19.4% and 25% correct, respectively, based on approximately 30 rounds of testing. More additional details were presented in Table S1 of SI.

**Differences found between ChatGPT's and Gemini's responses.**

Upon examination of ChatGPT and Gemini's thinking process based on the logs, here are some significant differences we found:

For this requirement: Horizontal Field Width (HFW) must be between 200 – 600 microns.
- ChatGPT: Adjusts HFW to 200 microns.
- Gemini: Adjusts HFW to 400 microns.

For this requirement: Description of the image
- ChatGPT: Gave more description on magnification and field width.

- Gemini: Gave more description on specific possible other phases present and regions they are in, though inaccurate.

*F. TEAMWORK PERFORMANCE*

After carrying out the collaborative experiment 20 times, 60-80% accuracy was achieved, with detailed elaborations under Table S2 and S3 in SI. This is a great improvement from the individual accuracies.

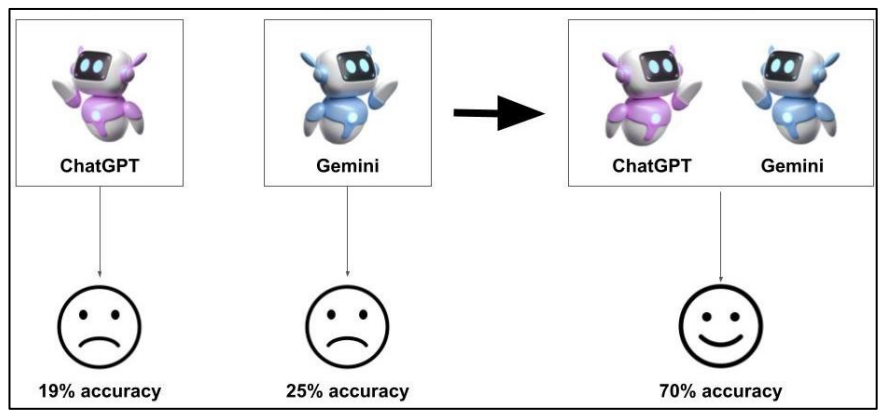

**Figure 3.** Performance improvement after teamwork

Initially, our objective was to conduct a comparative analysis of ChatGPT and Gemini to benchmark and determine their individual performance capabilities. Some benchmark criteria we used were number of function calls to determine efficiency, and of course accuracy of results. However, the results were unexpectedly underwhelming. The individual performance analysis revealed that ChatGPT attained an accuracy of 19%, while Gemini demonstrated a slightly higher accuracy of 25%, highlighting the challenges of single-model image analysis in this context. Despite our diligent efforts in refining and optimizing the prompts through multiple rounds of prompt engineering, we were unable to significantly enhance the accuracy levels of either system. This outcome underscores the challenges and complexities inherent in achieving high performance in such advanced AI models that operate independently.

Following the initial underperformance, we pivoted to a collaborative approach, allowing ChatGPT and Gemini to work in tandem for our qualitative task. We structured this by having ChatGPT review and provide feedback on Gemini's responses, then Gemini will receive the feedback and change its analysis accordingly if it agrees, engaging them in a constructive back-and-forth debate. This interaction was capped at a maximum of five exchanges per image analysis to maintain efficiency.

**EXPERIMENT II**

For proving generalizability, we conducted another experiment to challenge the LLMs to count the number of particles of a certain area based on the SEM images. An additional tool we used for this experiment is ImageJ, an image processing and analysis tool. The details on how the tool was used to generate the correct answer is further elaborated in Section VI.

The prompts we used are as follows:
Here are the prompts used:
ChatGPT Round 1:
Tell me how many white particles are larger than 10 micrometers^2 in this photo. Use appropriate techniques to isolate the white particles and exclude irrelevant regions like the scale bar. Ensure particles at the bottom that may be intersecting are not included. Use the scale bar at the bottom for pixel to micrometer conversion and show me an annotated image that highlights the detected particles, along with the number. The detection should focus on particles over 10 micrometers^2 and avoid any false positives from the scale bar region.

Gemini:
Assume the role where you are talking to ChatGPT, so your answer needs to be directly addressed to ChatGPT. I'm going to tell you what prompt i gave ChatGPT and what it responded me with. Evaluate its answer and give it feedback so that it can improve. You can also improve the prompt to help ChatGPT give a better answer.

Prompt for ChatGPT:
Tell me how many white particles are larger than 10 micrometers^2 in this photo. Use appropriate techniques to isolate the white particles and exclude irrelevant regions like the scale bar. Ensure particles at the bottom that may be intersecting are not included. Use the scale bar at the bottom for pixel to micrometer conversion and show me an annotated image that highlights the detected particles, along with the number. The detection should focus on particles over 10 micrometers^2 and avoid any false positives from the scale bar region.

ChatGPT's response: <ChatGPT Round 1's Response>

ChatGPT Round 2:
See the feedback below and try the analysis again:

<Gemini's Response>

A. Original Image

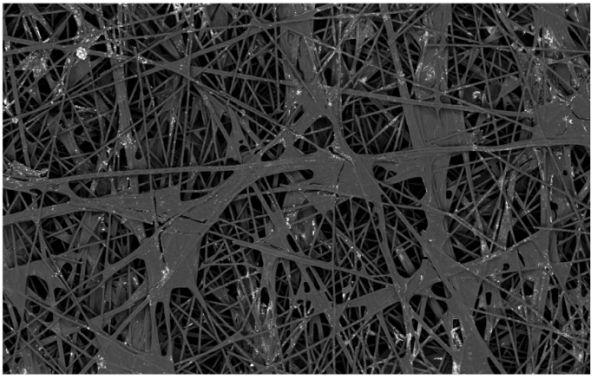

B. Particles identified by ImageJ

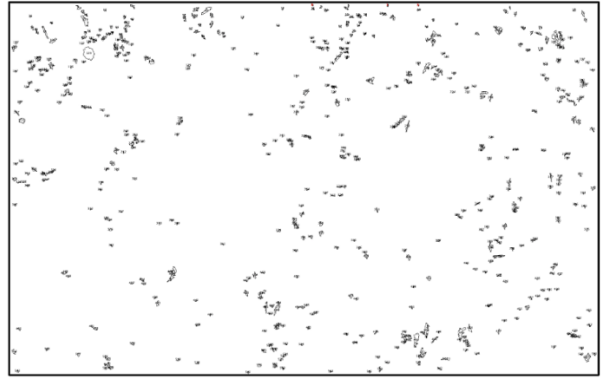

C. Before Gemini's Correction

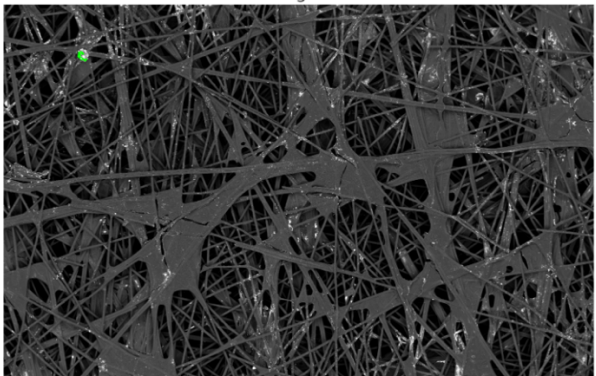

D. After Gemini's Correction

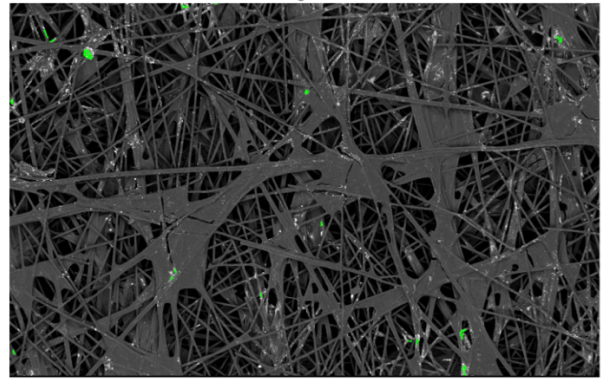

300 µm

**Figure 4.** Representative examples of particle counting based on one SEM image of the sample with metallic particles grown on the carbon network: (A) displays the original image we will process with both ChatGPT and ImageJ. (B) shows all particles identified by ImageJ that meets the objective criteria of having a particle size larger than 10 micrometers^2. (C) is the image produced by ChatGPT that identifies the particles in green. (D) is the image produced by ChatGPT after following Gemini's advice. The scale bar in the bottom left applies to all panels (A-D).

| Image | ChatGPT's First Answer | ChatGPT's Revised Answer | Improved? | Correct Answer |
|---|---|---|---|---|
| 1 | 6 | 5 | No | 517 |
| 2 | 253 | 271 | Yes | 432 |
| 3 | 1 | 338 | Yes | 669 |
| 4 | 4 | 16 | Yes | 706 |
| 5 | 17 | 31 | Yes | 578 |
| 6 | 0 | 3 | Yes | 321 |
| 7 | 4 | 67 | Yes | 546 |
| 8 | 1 | 16 | Yes | 359 |
| 9 | 0 | 2 | Yes | 459 |
| 10 | 44 | 44 | No | 628 |

**Table 2.** Overall results for Experiment II

Detailed responses and images can be found in Table S4, showcasing the responses for every prompt given, including the images produced - EXPERIMENT II: COUNTING PARTICLES in SI.

## III. DISCUSSION

In Experiment I, implementing a collaborative strategy based on structured debates between ChatGPT and Gemini led to substantial improvements, achieving an accuracy range of 60-80%. This experiment demonstrated that the debate mechanism effectively refines model outputs by enabling each AI system to critically evaluate and augment the other's insights. The result is a more accurate and reliable outcome than what either model could achieve independently, especially for complex image analysis tasks in materials phase analysis.

For Experiment II, we extended this approach to a quantitative task - particle counting - to test the adaptability of this collaborative framework. Here, Gemini's feedback successfully guided ChatGPT toward the correct answer 80% of the multiple times, underscoring the practical benefits of integrating feedback loops within an AI system. Even in numerical assessments, which traditionally rely on precise calculations rather than subjective reasoning, the synergy between the two models facilitated improved accuracy, confirming the value of using multiple LLMs for tasks that demand both qualitative and quantitative rigor.

This collaborative dynamic between ChatGPT and Gemini highlights the potential of leveraging multiple AI capabilities to enhance overall performance across a range of scientific tasks. By integrating diverse AI perspectives, this multi-agent system optimizes experimental outcomes and serves as a foundation for developing a modern, collaborative AI strategy that accelerates research in materials science. Such an approach has broader implications, suggesting that a similar framework could be beneficial in other scientific disciplines, where collaborative AI models can support rigorous, complex workflows. This paves the way for a future where AI not only assists but actively elevates research, promoting rapid advancements and a deeper understanding across various applications.

## DATA AVAILABILITY

Data will be made available on request.

## ACKNOWLEDGEMENT

We acknowledge support by the Defense Advanced Research Projects Agency (DARPA) under Agreement No. HR00112490369 and by DTRA (Award No. HDTRA1-20-2-0002) Interaction of Ionizing Radiation with Matter (IIRM) University Research Alliance (URA).

## REFERENCES

[1]  Wenya Linda Bi, Ahmed Hosny, Matthew B Schabath, Maryellen L Giger, Nicolai J Birkbak, Alireza Mehrtash, Tavis Allison, Omar Arnaout, Christopher Abbosh, Ian F Dunn, et al., "Artificial intelligence in cancer imaging: clinical challenges and applications", CA: a cancer journal for clinicians, vol. 69, no. 2, pp. 127–157, 2019.

[2]    Yilun Du, Shuang Li, Antonio Torralba, Joshua B Tenenbaum, and Igor Mordatch, "Improving factuality and reasoning in language models through multiagent debate", arXiv preprint arXiv: 2305.14325, 2023.

[3]    Anurag Garikipati, Jenish Maharjan, Navan Preet Singh, Leo Cyrus, Mayank Sharma, Madalina Ciobanu, Gina Barnes, Qingqing Mao, and Ritankar Das, "OpenMedLM: Prompt engineering can out-perform fine-tuning in medical question-answering with open-source large lan- guage models", AAAI 2024 Spring Symposium on Clinical Foundation Models.

[4]    Erin Hennessey, Matthew DiFazio, Ryan Hennessey, and Nicky Cas- sel, "Artificial intelligence in veterinary diagnostic imaging: A literature review", Veterinary Radiology & Ultrasound, vol. 63, pp. 851–870, 2022.

[5]    Ahmed Hosny, Chintan Parmar, John Quackenbush, Lawrence H Schwartz, and Hugo JWL Aerts, "Artificial intelligence in radiology", Nature Reviews Cancer, vol. 18, no. 8, pp. 500–510, 2018.

[6]    Mohamed R Ibrahim, James Haworth, and Tao Cheng, "Under- standing cities with machine eyes: A review of deep computer vision in urban analytics", Cities, vol. 96, pp. 102481, 2020.

[7]    Zhichu Ren, Zhen Zhang, Yunsheng Tian, and Ju Li, "Crest–copilot for real-world experimental scientist", DOI: 10.26434/chemrxiv-2023-tnz1x, 2023.

[8]    Jiaqi Wang, Zhengliang Liu, Lin Zhao, Zihao Wu, Chong Ma, Sigang Yu, Haixing Dai, Qiushi Yang, Yiheng Liu, Songyao Zhang, et al, "Review of large vision models and visual prompt engineering", Meta-Radiology, vol. 1, n. 3, pp. 100047, 2023